\documentclass[french]{pfia}

\usepackage{xcolor}
\definecolor{afiablue}{RGB}{61,159,207}
\definecolor{afiared}{RGB}{167,75,68}
\definecolor{afialightblue}{RGB}{158,193,232}
\usepackage{enumitem}
\setlist[itemize]{label=$\bullet$}
\addto\captionsfrench{}
\NoAutoSpaceBeforeFDP

\usepackage{graphicx}
\usepackage{float}
\usepackage{url}
\newcommand{\field}[1]{\textcolor{afiablue}{#1}}
\newcommand{\val}[1]{\textsf{\textcolor{afialightblue}{#1}}}
\usepackage{algorithm}
\usepackage{algpseudocode}
\usepackage{amsmath} 
\usepackage{tcolorbox}
\tcbset{colback=gray!10, colframe=black, boxrule=0.5pt}
\usepackage{cuted}
\usepackage{caption}
\usepackage{tabularx}
\usepackage{booktabs}
\usepackage{ragged2e}

\newcounter{prompt}
\newcommand{\promptcaption}[1]{%
  \refstepcounter{prompt}%
  \captionsetup{type=figure}%
  \vspace{-0.5em}%
  {\centering\caption*{Prompt \theprompt: #1}\par}%
}

\title{\textbf{Optimizing Hypergraph-Based RAG: Toward Better Fact Extraction and Chunk Retrieval}}

\author{Houda Khrouf, Pedro Fillastre, Sebastiao Correia \\[6pt]
Applied Research, Qlik \\
\texttt{\{houda.khrouf, pedro.fillastre, sebastiao.correia\}@qlik.com}}
\date{}

\begin{document}

\maketitle


\begin{abstract}
GraphRAG enables deeper reasoning by structuring knowledge as graphs but struggles with n-ary facts. HyperGraphRAG uses hypergraphs for richer semantics, improving accuracy, yet relies on error-prone LLM extraction and inefficient standard chunk retrieval. We address this by employing self-consistency prompting to improve the extraction, and Personalized PageRank algorithm over hypergraph to enhance chunk retrieval.
\end{abstract}



\section{Introduction}
Integrating external knowledge into large language models (LLMs) via Retrieval-Augmented Generation (RAG) has established itself as a key strategy for improving factual accuracy and reducing hallucinations \cite{lewis2020rag}. Standard RAG systems, based on vector similarity search across text segments (chunks), favor shallow semantic similarity at the expense of complex inter-entity relationships. While effective for targeted factual queries, they lack the contextual depth required for problems demanding cross-document understanding.

\begin{figure}[ht]
    \centering
\includegraphics[width=0.5\textwidth]{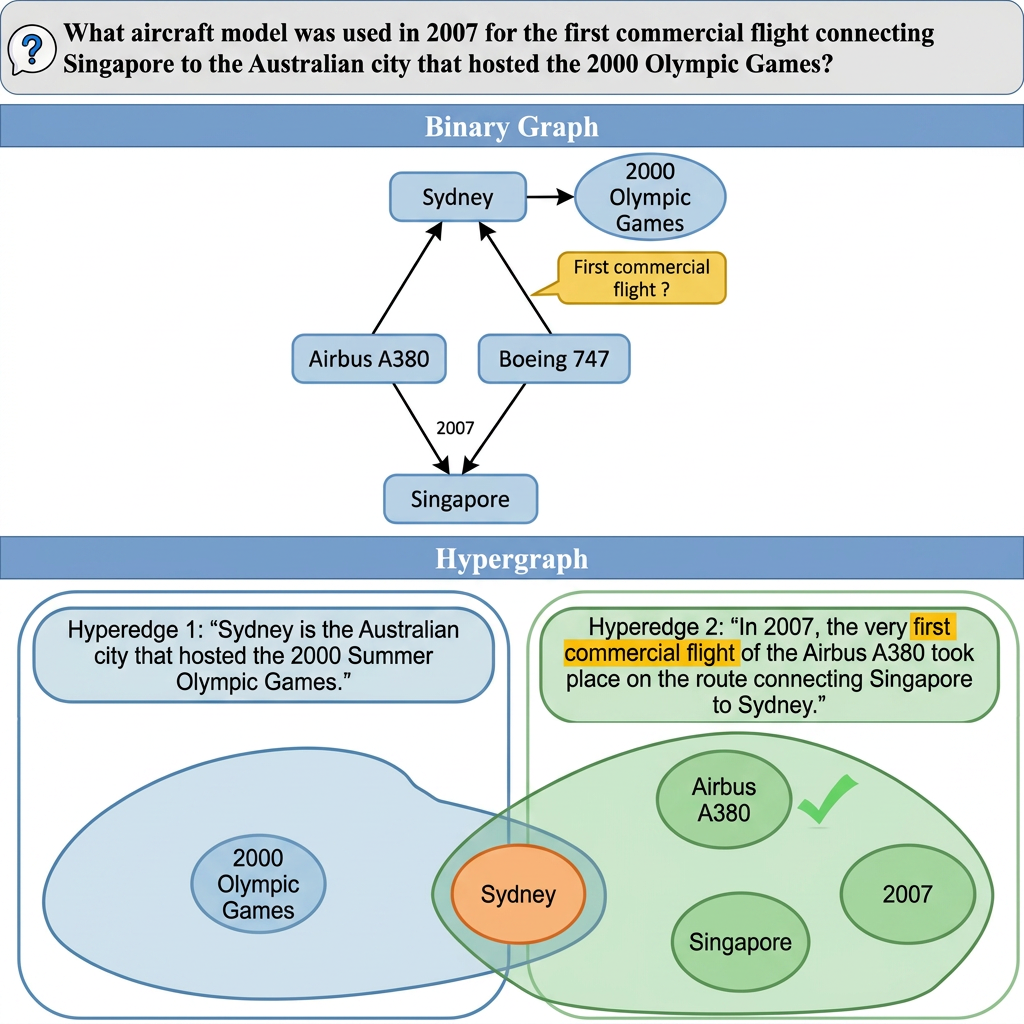}
    \caption{Comparison between a binary Knowledge Graph (KG) and a Knowledge HyperGraph (KHG) for answering a multi-hop question}
    \label{fig:hypergraph_example}
\end{figure}

To address these shortcomings, GraphRAG \cite{zulun2026survey, guo2024lightrag, chen2025pathrag} has emerged as a strategic evolution, structuring data in Knowledge Graphs. By modeling dependencies between nodes, GraphRAG enables finer-grained contextual understanding and efficient navigation through relational paths. This structure facilitates not only multi-hop reasoning and entity disambiguation, but also the handling of complex global queries (e.g., ``What are the main themes?'') that traditional RAG systems struggle to address. However, despite their ability to capture binary relations, classical knowledge graphs reach their limits when facing the intrinsic complexity of real-world data. Many relations are inherently $n$-ary, involving multiple entities simultaneously (for example, a collaboration among three companies or an event linking several actors to a specific location). Decomposing such relations into simple binary edges inevitably leads to a loss of semantic and structural information. To overcome this loss, approaches \cite{luo2025hypergraphrag, zai2026proh, feng2025hyperrag} based on knowledge hypergraphs have been proposed. Unlike binary graphs, hypergraphs use hyperedges to represent facts connecting multiple nodes simultaneously, thus offering superior expressiveness and faithful preservation of data integrity. As illustrated in Figure~\ref{fig:hypergraph_example}, the relation describing ``In 2007, the first commercial flight of the Airbus A380 connected Singapore to Sydney'' links four entities. In a binary graph, this relation is fragmented into independent links, making it impossible to distinguish it from other flights operated by different aircraft (such as a Boeing 747) sharing certain attributes (departure, destination, date).

However, despite their superior expressiveness, hypergraph-based approaches present two major limitations:

\begin{itemize}
\item \underline{Complexity of hypergraph construction}: Although hyperedge-based representation better preserves factual integrity than binary graphs --- which are often sensitive to linguistic complexity (verb tenses, negations, etc.) \cite{Xu2023LargeLM} --- its LLM-based extraction remains subject to structural instabilities: omitted entities or isolated hyperedges, and deficient coreference resolution hindering data continuity. These shortcomings fragment the graph topology and limit the model's ability to navigate across knowledge, directly impacting multi-hop reasoning.

\item \underline{Under-exploitation of structural connectivity}: The retrieval strategy employed in HyperGraphRAG \cite{luo2025hypergraphrag} relies on semantic search combined with local one-hop neighborhood expansion. This mechanism under-exploits the global topology of the hypergraph and suffers from a horizon problem: depending on the semantic scope of the query, it cannot capture structurally connected fragments beyond the immediate neighborhood. This approach overlooks fragments with significant structural dependencies. It misses the opportunity to prioritize passages by their degree of connectivity with relevant entities, limiting multi-hop reasoning.
\end{itemize}

To address these limitations, we propose two contributions. First, an optimized extraction method (EXT\textsuperscript{++}) based on \textit{self-consistency prompting}, which improves the completeness and connectivity of the hypergraph without additional extraction cost. Second, a retrieval mechanism based on Personalized PageRank (PPR) operating on the hypergraph, inspired by HippoRAG2 \cite{gutierrez2025hipporag2}, which identifies the most relevant chunks through their structural connectivity rather than through simple vector similarity. The source code and evaluation data are publicly available\footnote{\url{https://github.com/qlik-oss/HyperGraphRAG/}} to ensure reproducibility.

\section{Related Work}
\label{sec:relatedwork}

The RAG architecture is built on a modular pipeline organized around three successive phases: pre-retrieval, retrieval, and generation. The initial pre-retrieval phase defines the representation and structure of source data through text segmentation and granularity selection (from sentence to document level), sometimes incorporating Knowledge Graph modeling to optimize semantic context. Text segmentation is essential to overcome token constraints of language models while improving the precision and efficiency of RAG system. Initial strategies, such as fixed-size or recursive chunking, are simple to implement but can compromise semantic coherence. This renders the system ineffective for answering global questions and exacerbates hallucinations on complex relations. To address this, contextual chunking \cite{anthropic2024contextual} enriches each chunk with a summary of its global context within the document, thereby limiting information loss during indexing and improving the retrieval of relevant chunks. Other knowledge-graph-based approaches transcend chunking limitations by interconnecting entities through semantic relations, shifting from flat textual similarity search to deep structural and relational understanding of data. At the core of this evolution, Microsoft's GraphRAG \cite{edge2024graphrag} structures the corpus as a knowledge graph and leverages community detection to produce hierarchical summaries, facilitating the processing of global queries. Other recent works have sought to improve the expressiveness of knowledge representation and the quality of graph-based reasoning. LightRAG \cite{guo2024lightrag} relies on incremental indexing and two-level retrieval: a fine-grained search focused on relevant entities and relations, combined with a higher-level thematic search. HippoRAG2 \cite{gutierrez2025hipporag2} proposes an architecture inspired by human associative memory and the hippocampus, using Personalized PageRank to identify relevant chunks by exploiting graph topology. PathRAG \cite{chen2025pathrag} refines retrieval through reasoning path pruning to select the most informative subgraphs while limiting the introduction of noise into the context provided to the model.

The evolution of these systems has led to a diversification of graph types for data modeling. Notable types include: (i) the classical knowledge graph (KG), which extracts binary relations between entities \cite{gutierrez2025hipporag2, zulun2026survey}, sometimes enriched with a taxonomic or ontological layer to improve semantic structuring and disambiguation \cite{lei2025kag}; (ii) the text graph (or attributed graph), where each node corresponds to a text segment with metadata enabling multi-hop retrieval via neural (GNN) or structural mechanisms \cite{he2024gretriever, hu-etal-2025-grag}; and (iii) the bipartite graph (passage--entity), which explicitly models co-occurrences to provide compact indexing for guiding contextual search \cite{yiqian2025ketrag}. Since these representations remain limited to binary relations, recent research has turned to hypergraphs for modeling the higher-order interactions discussed in the introduction. In this regard, HyperGraphRAG \cite{luo2025hypergraphrag} lays the foundations of this paradigm by leveraging hyperedges to represent complex facts. This approach surpasses classical graphs through a more faithful capture of entity co-occurrences, thus offering superior structural expressiveness. In parallel, Hyper-RAG \cite{feng2025hyperrag} proposes a hybrid architecture. By jointly extracting binary and $N$-ary relations coupled with vector indexing, it provides the LLM with factual context of unprecedented density during the retrieval and generation phases, drastically reducing the risk of hallucinations. More recently, PRoH \cite{zai2026proh} marks a further step by focusing on adaptive reasoning planning. The system decomposes queries into a directed acyclic graph (DAG) of sub-questions and iteratively navigates the local neighborhood of the hypergraph. Although generation quality is notably improved, the impact of this iterative process on system latency remains a blind spot in its evaluation. These works leave open questions regarding the reliability of large-scale extraction and the exploitation of the global topology of the hypergraph during retrieval.

\section{Methodology}
The proposed methodology is organized around two complementary axes aimed at improving the reliability of hypergraph construction and optimizing the retrieval of relevant chunks.

\subsection{Optimized Extraction (\texorpdfstring{$EXT^{++}$}{Ext++})}
LLM-based knowledge graph extraction is increasingly moving toward the Open Information Extraction (OpenIE) paradigm. Freeing themselves from \textit{a priori} ontologies, LLMs are leveraged to dynamically discover and extract latent entities and relations directly from document corpora (bottom-up discovery). However, since LLM architectures are fundamentally optimized for sequential token generation rather than structured information extraction, this generative nature raises significant challenges. Beyond semantic proliferation, the extraction process encounters a granularity gap between the fluidity of natural language and the rigidity of the triplet format. It primarily suffers from structural failures, such as incompleteness when dealing with complex action verbs or graph fragmentation due to imperfect coreference resolution. To these are added cognitive biases of the model, notably semantic drift: by oversimplifying relations, the system transforms nuances (uncertainties, hypotheses, temporal aspects) into established facts. This inability to model negation or doubt generates factual ``noise,'' degrading the fidelity of the graph \cite{Xu2023LargeLM}. Hypergraph-based extraction intrinsically mitigates these cognitive failures by preserving the semantic integrity of facts. However, it remains vulnerable to structural instabilities: omission of entities, production of isolated hyperedges, sometimes devoid of genuine semantic meaning (e.g., residual fragments of the ``Who am I?'' type).

Although fine-tuning on graph extraction tasks constitutes a promising avenue \cite{Xu2023LargeLM}, its application remains conditioned on the availability of labeled corpora, which are currently lacking for hypergraphs. We therefore oriented our approach toward optimization through advanced prompt engineering. We propose an extension of HyperGraphRAG's few-shot extraction \cite{luo2025hypergraphrag} by integrating a self-consistency mechanism: \textit{self-consistency prompting}. This mechanism is inspired by the Universal Self-Consistency (USC) paradigm \cite{chen2024universal}, a method that enables LLMs to evaluate and aggregate multiple generation paths to produce a robust consensus without an external evaluator. This integration offers three fundamental advantages. First, it reduces hallucinations and semantic drift \cite{wang2023selfcons}: by relying on the convergence of multiple extractions rather than a single greedy decoding, erroneous or speculative relations are statistically filtered. Second, it mitigates the inherent variance of LLMs and their sensitivity to positional bias in long contexts, a frequent problem leading to the omission of peripheral entities. Finally, union-based aggregation across different iterations maximizes the completeness of the hypergraph.

Concretely, our EXT\textsuperscript{++} method relies on a prompt presented in Appendix~\ref{appendix:prompt}. This prompt executes three extraction iterations within a single LLM call for a given chunk, internally merged by union to ensure a denser and more connected final topology. Beyond this aggregation, EXT\textsuperscript{++} rethinks the instruction structure within the prompt: the model is explicitly constrained to list the concerned entities immediately after the definition of each hyperedge, following their order of appearance in the source text. Additionally, a coreference resolution instruction is applied during extraction (e.g., substituting a pronoun with the full named entity). This targeted extraction engineering enables (1) improved traceability of extracted information, (2) prevention of entity omission, (3) facilitation of the unambiguous topological association between a hyperedge and its entities, and (4) minimization of errors related to pronominal ambiguities.

\subsection{Optimized Retrieval via Personalized PageRank (PPR) on Hypergraph}
In the HyperGraphRAG approach \cite{luo2025hypergraphrag}, retrieval relies on semantic search over entities and hyperedges followed by a bidirectional one-hop topological expansion, supplemented by chunks from a standard vector search. As discussed in the introduction, this strategy under-exploits the global graph topology, and dense vector search tends to introduce contextual noise --- redundant or structurally disconnected passages from the actual need --- which dilutes useful signals and increases the model's vulnerability to hallucinations.

To address these limitations, we directly integrate chunks as nodes in the hypergraph, explicitly linking them to the entities and hyperedges derived from them. On this unified structure, we apply the Personalized PageRank (PPR) algorithm, inspired by HippoRAG2 \cite{gutierrez2025hipporag2}, treating entities, hyperedges, and chunks as nodes of a tripartite graph. The objective is to identify chunks that are both semantically relevant and strongly connected, topologically, to the query content. PPR acts as a robust filter against false positives from dense search, enabling refinement of the extracted subgraph quality before response generation.

\begin{figure}[ht]
    \centering
\includegraphics[width=0.5\textwidth]{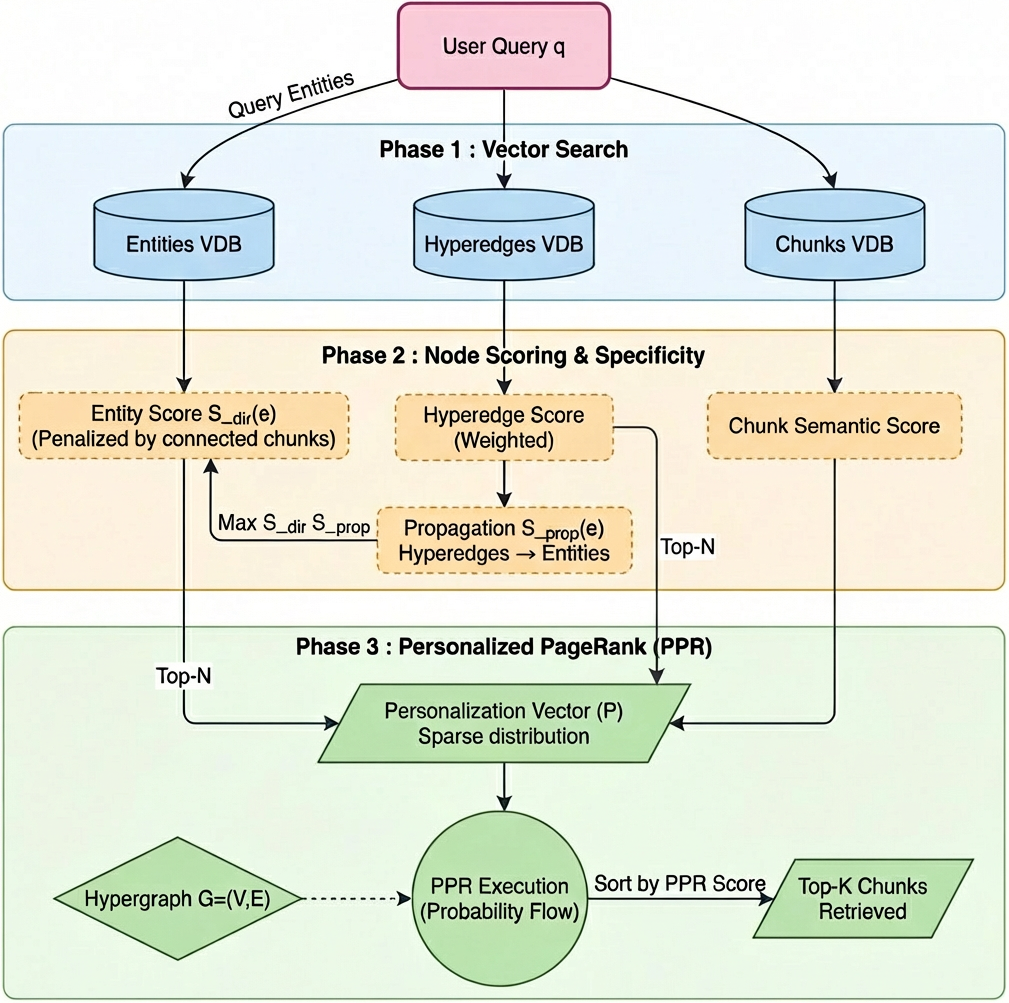}
    \caption{Retrieval pipeline based on Hypergraphs and Personalized PageRank (PPR)}
    \label{fig:workflow_ppr}
\end{figure}

As illustrated in Figure~\ref{fig:workflow_ppr}, the optimized retrieval consists of three steps:

\vspace{2mm}
\textbf{Step 1: Vector Search.} The user query $q$ is used to perform a vector search over the hyperedge and chunk databases, while entities extracted from $q$ are used to query the entity database. Retrieved nodes are sorted by decreasing similarity, and only the top $k$ of each type are retained. However, hyperedges undergo a specific sorting based on a hybrid criterion combining their similarity and their importance weight in the source document, assigned during extraction.

\vspace{2mm}
\textbf{Step 2: Node Scoring.} Each retrieved element (entity, hyperedge, chunk) receives an initial relevance score corresponding to its similarity with the query. Specific weighting strategies are applied to calibrate these scores. On the one hand, inspired by the established effectiveness of global importance signals such as IDF (Inverse Document Frequency) for information retrieval, a specificity penalty is applied to entities: the score is divided by the degree of connectivity to chunks ($|C_e|$) to penalize generic entities:

\begin{equation}
s_{\text{local}}(e) = \frac{\text{sim}(e, q)}{|C_e|}
\end{equation}

A key mechanism in this step is the relevance propagation ($S_{prop}$) from hyperedges to associated entities. This enriches the PPR personalization vector with entities that would not have been identified through vector search on entity names alone, broadening the coverage of the query signal within the graph. The propagated score of an entity $e$ is computed as the average of the scores of incident hyperedges, weighted by connectivity:

\begin{equation}
\bar{s}(e) = \frac{1}{|\mathcal{H}_e|} \sum_{h \in \mathcal{H}_e} \frac{\mathrm{sim}(h, q)}{|C_e|}
\end{equation}

Then, a Noisy-OR aggregation amplifies the score of entities connected to multiple hyperedges while keeping it bounded in $[0, 1]$:

\begin{equation}
s_{\text{prop}}(e) = 1 - \left(1 - \bar{s}(e)\right)^{1 + \ln|\mathcal{H}_e|}
\end{equation}

where $\mathcal{H}_e$ denotes the set of retrieved hyperedges incident to $e$, $\mathrm{sim}(h, q)$ the vector similarity between hyperedge $h$ and query $q$, and $|C_e|$ the number of chunks associated with $e$. The intuition behind Noisy-OR is that each incident hyperedge constitutes an independent observation: the more an entity is linked to relevant hyperedges, the higher its score. The final score of each entity is then determined by:

\begin{equation}
s(e) = \max\!\left( s_{\text{local}}(e),\; s_{\text{prop}}(e) \right)
\end{equation}

On the other hand, the chunk score corresponds to their similarity modulated by a weighting factor $w_{chunk}$, balancing their influence relative to conceptual nodes (entities and hyperedges) in the PPR algorithm.

\vspace{2mm}

\textbf{Step 3: Personalized PageRank.} The computed scores are Min-Max normalized and then assembled into a sparse personalization vector merging the Top-N entities and hyperedges, along with the weighted similarity scores of chunks. This vector encodes the query bias over all nodes of the hypergraph. The PPR algorithm propagates this signal through the structure, redistributing the importance of each node based on its structural proximity to the relevant elements. Chunk-type nodes are then extracted and sorted by decreasing PPR score, yielding the Top-K most relevant passages for generation.

\section{Experiments}

\begin{table*}[t]
    \centering
    \includegraphics[width=\textwidth]{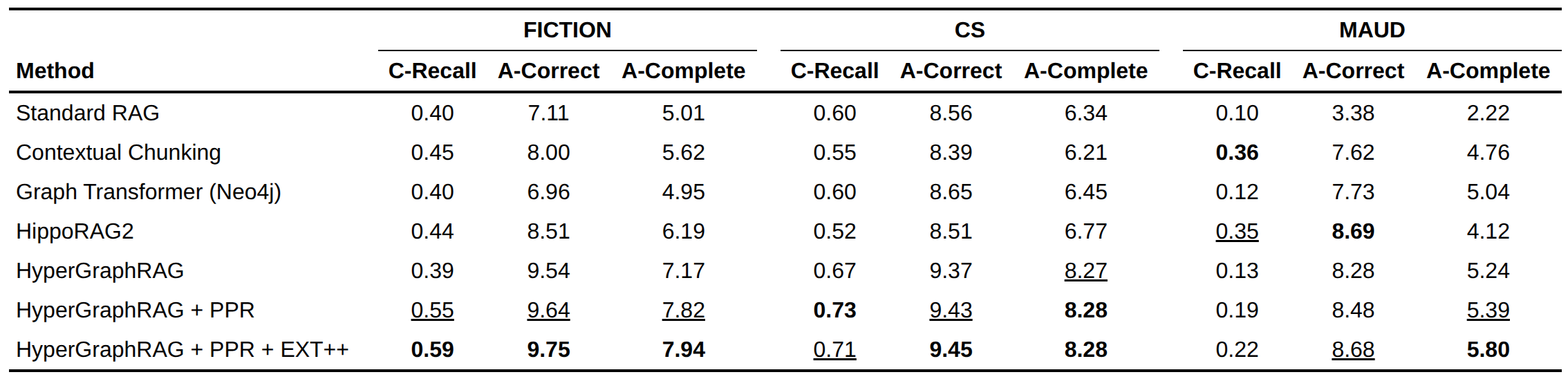} 
    \caption{Performance comparison across different domains. Best scores are in bold and second-best are underlined for each dataset.}
    \label{tab:baselines_table}
\end{table*}

\subsection{Evaluation Framework}

\paragraph{Datasets.} To evaluate the effectiveness of our optimization strategy, we selected three datasets from different benchmarks presenting distinct challenges. Particular attention was paid to excluding Wikipedia-based corpora, which are ubiquitous in LLM training data, to prevent memorization bias.

\begin{itemize}[leftmargin=1.1em]
  \item \textbf{Fiction}: Drawn from the UltraDomain benchmark \cite{qian2025memorag}, it evaluates RAG systems on the analysis of long and complex narratives. Due to the extreme length of documents --- reaching up to 560k tokens and requiring a high number of LLM calls per document for hypergraph extraction --- we constructed a reduced version by randomly selecting 13 out of 30 documents to control experimentation cost and time. This subset comprises 84 questions, of which 72 are multi-hop queries requiring the synthesis of 2 to 10 chunks. These questions cover five analytical dimensions: thematic content, character development, plot, literary techniques, and meta-fictional context.
  \item \textbf{CS} (Computer Science): Drawn from the HyperGraphRAG paper \cite{luo2025hypergraphrag} and originally derived from the UltraDomain benchmark \cite{qian2025memorag}, it is based on voluminous university textbooks specialized in computer architecture, mathematical algorithms, and machine learning. It contains 3 dense technical documents with 398 questions, of which 175 are multi-hop queries. These questions are constructed from knowledge fragments located at a distance of 1 to 3 hops, involving fact retrieval and multi-entity synthesis tasks.
  \item \textbf{MAUD} (Merger \& Acquisition Understanding Dataset): Drawn from the LegalBench-RAG benchmark \cite{pipitone2024legalbenchrag}, it is dedicated to the analysis of complex contractual clauses within merger and acquisition agreements. Identified as the most demanding component of the benchmark due to its highly specialized legal and financial terminology, it consists of single-hop factual questions, meaning the answer can be extracted directly from a single passage. For our work, we constructed a reduced and representative version by randomly selecting 21 out of 150 available documents. This subset comprises 74 questions and helps control experimentation cost and time.
\end{itemize}

\paragraph{Baselines.} To evaluate the performance of our approach, we compared it against several state-of-the-art methods, organized into four distinct categories:

(i) \textbf{Standard RAG}: the conventional approach based on chunk segmentation and leveraging dense or hybrid vector similarity search (combining semantic and lexical search); (ii) \textbf{RAG with enriched segmentation}, exploring advanced chunking strategies. Although various methods were evaluated on LegalBench-RAG \cite{pipitone2024legalbenchrag} (including semantic chunking, proposition chunking, and summary chunking), Anthropic's contextual chunking \cite{anthropic2024contextual} proved to be the best-performing in our preliminary tests and was retained as a baseline; (iii) \textbf{Graph-based RAG}, where two baselines are retained: HippoRAG2 \cite{gutierrez2025hipporag2}, described in Section~\ref{sec:relatedwork}, and GraphTransformer \cite{neo4j_graphtransformer}, which relies on a binary entity graph representation and benefits from native integration with the LangChain and Neo4j libraries. During retrieval, this approach identifies query entities and traverses the graph to extract relevant neighborhoods; (iv) \textbf{Hypergraph-based RAG}: To model higher-order relations, we integrated HyperGraphRAG \cite{luo2025hypergraphrag}, which we extend in this work. Evaluated without our optimization, it serves as a baseline to measure the contribution of our approach.

\paragraph{Metrics.}
Our evaluation relies on three main metrics: Contextual Recall, Correctness, and Completeness. Contextual Recall, implemented in the RAGAS library \cite{es-etal-2024-ragas}, measures the proportion of relevant information extracted from reference documents that is effectively reused in the generated response. In other words, it evaluates the system's ability to retrieve and mobilize the necessary context to produce a relevant answer.

Furthermore, to evaluate correctness and completeness, we adopt an LLM-as-a-judge approach using the GPT-4.1-mini model. This approach compares the generated response against a reference answer using a prompt inspired by the one defined in \cite{luo2025hypergraphrag}. Correctness evaluates the extent to which the response is logically and factually aligned with the reference, while completeness evaluates whether it covers all essential aspects. Both dimensions are scored on a scale of 0 to 10, following well-defined criteria in the prompt presented in Appendix~\ref{appendix:prompt}.

We chose to exclude the F1 score from our evaluation, despite its widespread use in the literature. This metric relies on lexical similarity between the generated response and the reference, making it particularly sensitive to variations in wording, length, and tokenization. In the RAG context, where the same question may admit multiple correct answers with different formulations, this dependence on lexical matching can introduce evaluation bias.

\paragraph{Implementation.}
The GPT-4o-mini model (distinct from the model used for evaluation) is used for hypergraph extraction and response generation, while \texttt{NovaSearch/stella\_en\_400M\_v5} is used for vector representation. The top-k parameter is set to 5 chunks for the CS and MAUD datasets, and to 10 for Fiction where some queries require more than 5 chunks. For the retrieval phase, we retain the original HyperGraphRAG parameters:
$k_V = 60$ candidate entities and $k_H = 60$ candidate hyperedges are extracted via vector search, then truncated according to a budget of 4{,}000 tokens each. For Personalized PageRank (PPR), we set its damping factor $\alpha = 0{.}5$ (value selected by HippoRAG2 following their hyperparameter tuning on a training data subset), the chunk weighting factor $w_{\text{chunk}} = 0{.}5$ (determined by empirical validation across the 3 datasets by maximizing recall), and we initialize the personalization vector on the top $k_{\text{ent}} = 5$ entities and $k_{\text{hyp}} = 10$ hyperedges. We use NetworkX for hypergraph management and NanoVectorDB for chunk vector storage and search. NanoVectorDB is a lightweight vector database designed for in-memory embedding storage and retrieval.

\subsection{Results}

\paragraph{Comparison with baselines:}
Table~\ref{tab:baselines_table} presents the results of our comparative evaluation on the three datasets (Fiction, CS, MAUD). Our approach (HyperGraphRAG + PPR + EXT++) achieves the best performance or ranks second on nearly all metrics. It achieves a remarkable gain in contextual recall (+51\% on Fiction and +69\% on MAUD), as well as in completeness, with +11\% on both corpora compared to HyperGraphRAG. These gains illustrate the complementarity of PPR and EXT++: the former improves the selection of relevant passages through probabilistic graph exploration, while the latter strengthens the quality of the underlying graph by reducing isolated and redundant hyperedges. Compared to Standard RAG, the gaps are even larger: completeness improves by +59\% on Fiction and +31\% on CS, and gains exceed +150\% on MAUD in both correctness and completeness, confirming the inability of vector search alone to handle multi-hop corpora (Fiction, CS), as well as specialized corpora like MAUD, where standardized terminology and the redundancy of contractual language make passage discrimination particularly difficult.

The results analysis  highlights the structural limitations of each baseline that our approach progressively resolves. Standard RAG, based on simple vector similarity between the query and chunks, has no explicit mechanism for contextual structuring or multi-hop reasoning. It fails particularly on MAUD (contextual recall=10\%, answer correctness=3.38), where many chunks are semantically very similar to each other, preventing dense search from discriminating the relevant passage among dozens of quasi-identical formulations, which explains the sharp drop in recall. Anthropic's Contextual Chunking partially mitigates this problem by enriching each chunk with a contextual summary of the document to which it belongs. This contextualization reinforces useful semantic signals and allows the dense search engine to better identify the relevant document. It proves particularly effective on MAUD, where the chunk context provides strong cues for identifying the correct contract among similarly worded documents --- for example, locating the right merger agreement containing the sought covenants. However, this strategy also introduces a side effect: the systematic addition of context can amplify certain lexical signals that are frequent but not discriminative in the corpus. When these signals dominate the embeddings, dense search can be diverted toward contextually close but irrelevant passages. This phenomenon is particularly visible in narrative corpora like Fiction, where contextual enrichment slightly improves retrieval but translates into only a limited gain in correctness (8.00), as contextual cues remain diffuse and difficult to leverage for precisely identifying target information. LangChain's Graph Transformer, which extracts a binary knowledge graph and retrieves one-hop (\emph{hop-1}) neighboring entities, suffers from a dual problem of fragmentation and noise. On the one hand, the binary graph is often highly fragmented, making context expansion difficult without introducing irrelevant information. On the other hand, there is an intrinsic trade-off between retrieval depth and relevance: low depth produces insufficient context, while high depth, without an intelligent pruning strategy, introduces considerable noise that degrades answer quality. HippoRAG2 partially addresses these limitations by applying Personalized PageRank on a binary graph, offering a more sophisticated probabilistic propagation mechanism than simple \emph{hop-1} traversal. This improves certain scores, but completeness remains low because binary triplets do not faithfully capture the n-ary, temporal, and modal relations frequent in the corpora.

Our approach resolves these limitations by extending HyperGraphRAG through the combination of three complementary mechanisms. First, the original hypergraph already offers a semantically richer representation than binary triplets and a naturally less fragmented graph --- which explains the performance jump of HyperGraphRAG compared to binary graph baselines~\cite{luo2025hypergraphrag}. However, without a global propagation mechanism, passage selection relies solely on local similarity between the query and graph entities, limiting contextual recall. The integration of PPR resolves this limitation by enabling probabilistic, weighted relevance propagation throughout the graph, avoiding the introduction of noise through score attenuation with topological distance. Finally, EXT\textsuperscript{++} intervenes upstream by improving the quality of the hypergraph itself, reducing the fraction of isolated hyperedges and strengthening graph connectivity. This translates into a slight gain in recall and response completeness. The synergy between these three components explains the superiority of our approach across all datasets. However, on MAUD, the contextual recall of our approach remains lower than that of HippoRAG2. The redundancy of contractual language penalizes hyperedges, which are richer in content than simple triplets: their formulations overlap more, introducing noise into the similarity signal during PPR initialization. HippoRAG2's binary triplets, being shorter and more discriminative, offer more precise anchoring in this type of corpus. Conversely, this richness explains the superiority in completeness (+41\% gain): the retrieved entities, hyperedges, and chunks provide the LLM with semantically denser context, enabling more exhaustive responses.

\paragraph{LLM model performance comparison:}

\begin{figure}[ht]
    \centering
\includegraphics[width=0.5\textwidth]{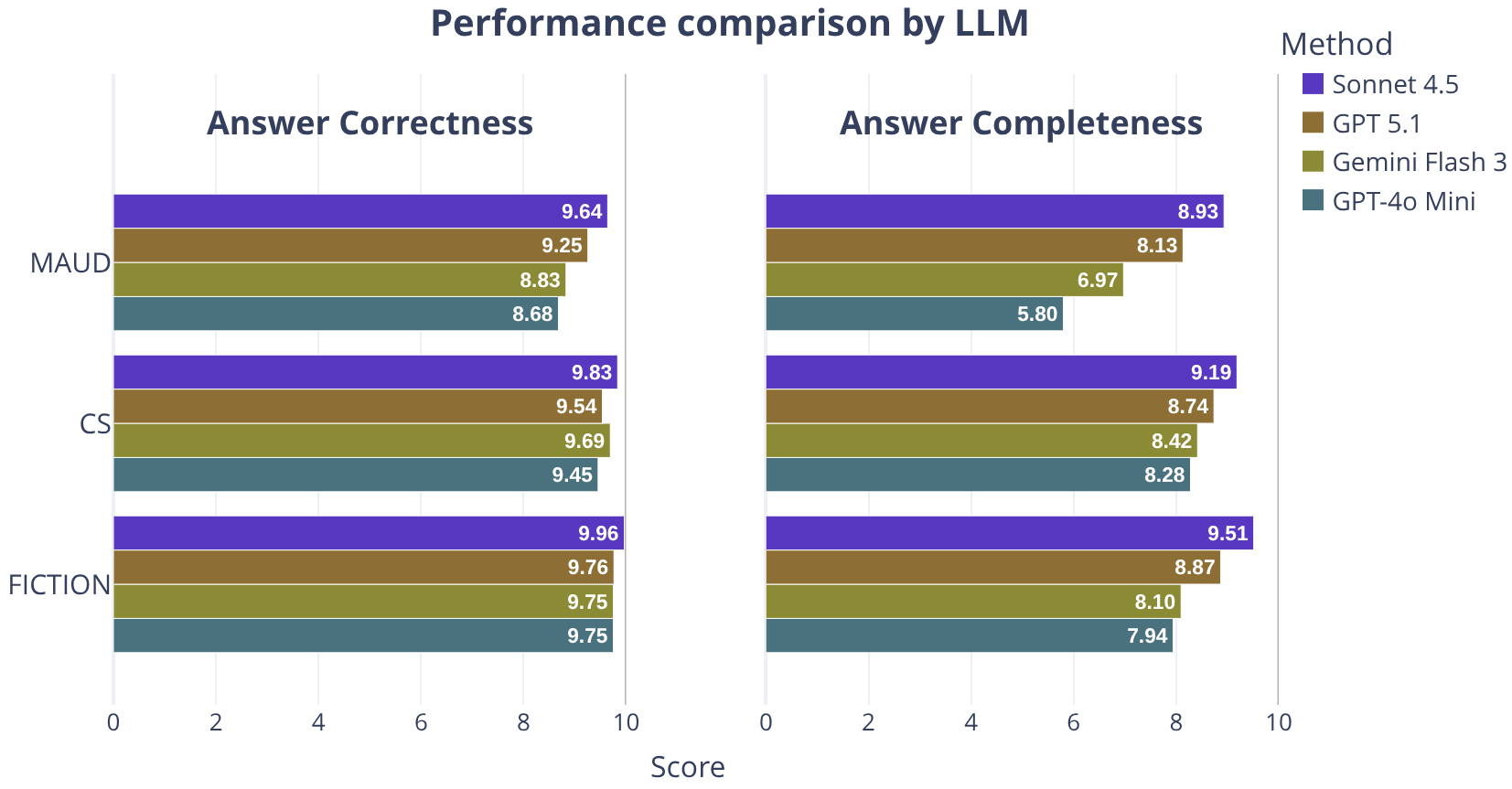}
    \caption{LLM model performance comparison for response generation}
    \label{fig:models_cmp}
\end{figure}

Figure~\ref{fig:models_cmp} and Table~\ref{tab:cost-comparison} respectively present the quality scores and the average cost per query obtained by four generation models (Claude Sonnet 4.5, GPT-5.1, Gemini Flash 3, and GPT-4o mini) used exclusively at the response generation step within our approach. Entity and relation extraction remains handled by GPT-4o mini for all experimental configurations. This architectural choice is motivated by economic constraints: extracting a hypergraph requires one LLM call per chunk in the corpus, for example over 1,500 calls for the MAUD dataset. Using an expensive model at this stage would incur prohibitive additional cost --- a factor of 25 between GPT-4o mini and Sonnet 4.5 --- making large-scale indexing economically unviable. Comparing models on response generation is important because they differ in their ability to identify and synthesize relevant information from the retrieved context containing noise.

\begin{table}[ht]
\centering
\begin{tabular}{|l|c|c|c|}
\hline
\textbf{Model} & \textbf{Fiction} & \textbf{MAUD} & \textbf{CS} \\
\hline
Sonnet 4.5 & \$0.0720 & \$0.0900 & \$0.0660 \\
GPT-5.1 & \$0.0261 & \$0.0360 & \$0.0214 \\
Gemini Flash 3 & \$0.0087 & \$0.0121 & \$0.0087 \\
GPT-4o mini & \$0.0028 & \$0.0036 & \$0.0026 \\
\hline
\end{tabular}
\caption{Average cost per query by model and use case}
\label{tab:cost-comparison}
\end{table}

The results reveal that \emph{correctness} scores remain high and relatively homogeneous across models (8.68 to 9.96 on a scale of 10), suggesting that the hypergraph-based retrieval architecture provides sufficiently relevant context for all evaluated models to produce factually correct responses. In contrast, the \emph{completeness} metric highlights more pronounced differences: Sonnet 4.5 achieves scores of 9.51 on Fiction and 8.93 on MAUD, while GPT-4o mini obtains 7.94 and 5.80, respectively, representing a degradation between 16\% and 35\%. This disparity indicates that more powerful models excel at discriminating relevant passages and exhaustively covering key information, whereas more compact models tend to omit certain elements in specialized domains.

This cost-quality trade-off provides operational flexibility depending on application requirements. For critical use cases such as legal analysis or due diligence, investing in a high-capacity reader model is justified, with the additional cost per query remaining moderate (\texttildelow\$0.07). For applications tolerating slightly reduced completeness or less demanding narrative domains, GPT-4o mini constitutes an economically viable alternative while maintaining a satisfactory level of correctness.

\begin{figure}[ht]
    \centering
\includegraphics[width=0.5\textwidth]{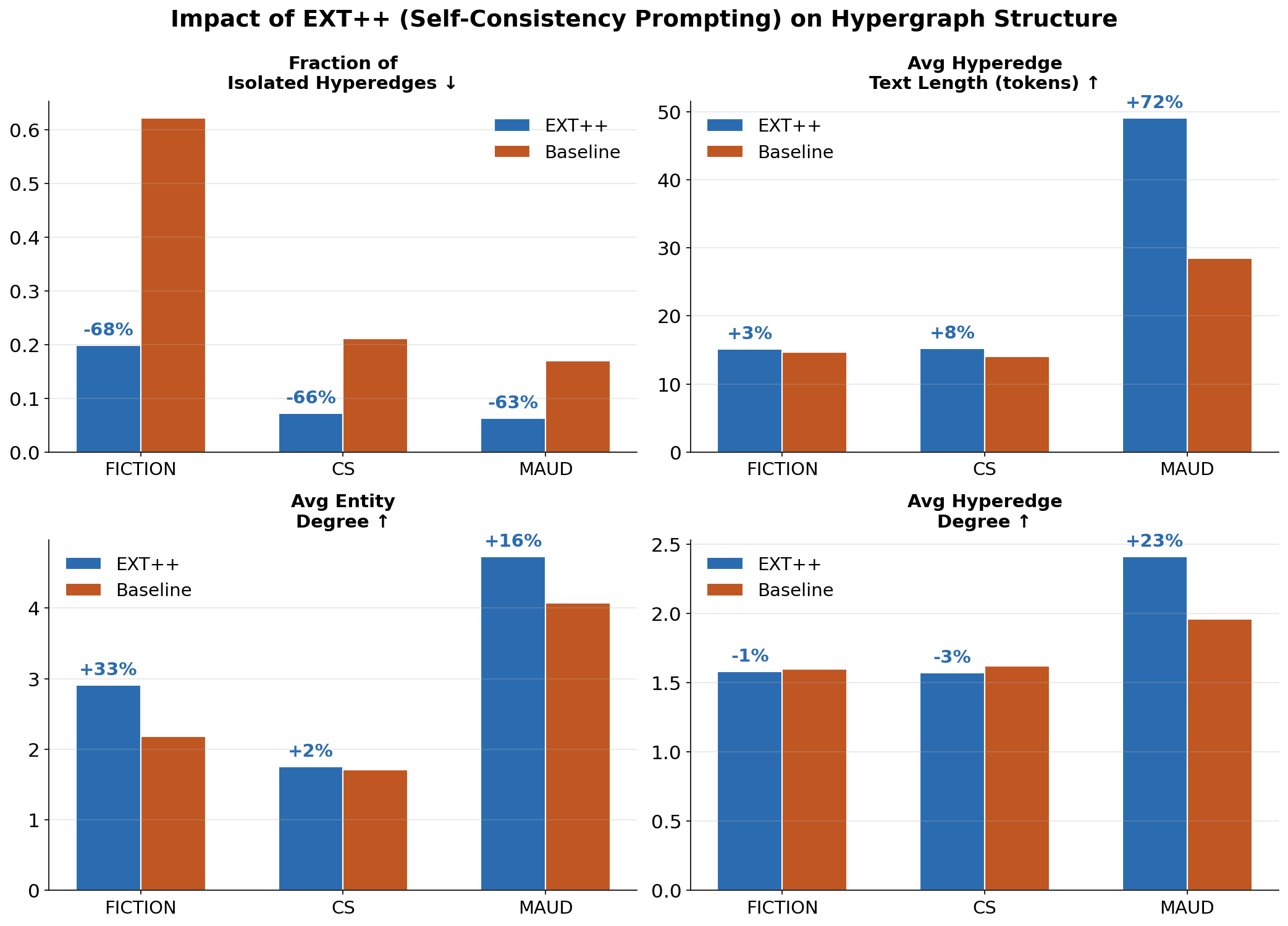}
    \caption{Hypergraph structure improvement through the EXT\textsuperscript{++} method}
    \label{fig:ext_benefit}
\end{figure}

\paragraph{Impact of EXT\textsuperscript{++} extraction:}
We examine the benefit of optimized EXT\textsuperscript{++} extraction on the structural quality of constructed hypergraphs. As illustrated in Figure~\ref{fig:ext_benefit}, EXT\textsuperscript{++} significantly reduces the fraction of isolated hyperedges across all three datasets, lowering the isolation rate from 62\% to 20\% on Fiction. Consequently, nearly all extracted facts are anchored to at least one entity, which strengthens the relational signal and ensures much better graph navigability during the search phase. Meanwhile, the increase in average entity degree reflects a densification of the semantic network; this topology favors multi-hop reasoning by multiplying information propagation paths. This optimization also produces longer and more informative hyperedge descriptions (for example, from 28 to 49 tokens for MAUD, a 72\% increase), thereby enriching the semantic representation. For MAUD, the average number of entities per connected hyperedge increases from 1.96 to 2.41 (+23\%), illustrating improved modeling of multi-entity relations, despite a negligible decrease (less than 3\%) on the other two datasets. In summary, EXT\textsuperscript{++} systematically produces more connected, more descriptive, and structurally richer hypergraphs, which directly translates into better coverage and improved response quality during reasoning.

\paragraph{Extraction cost and latency:}

\begin{figure}[ht]
    \centering
\includegraphics[width=0.5\textwidth]{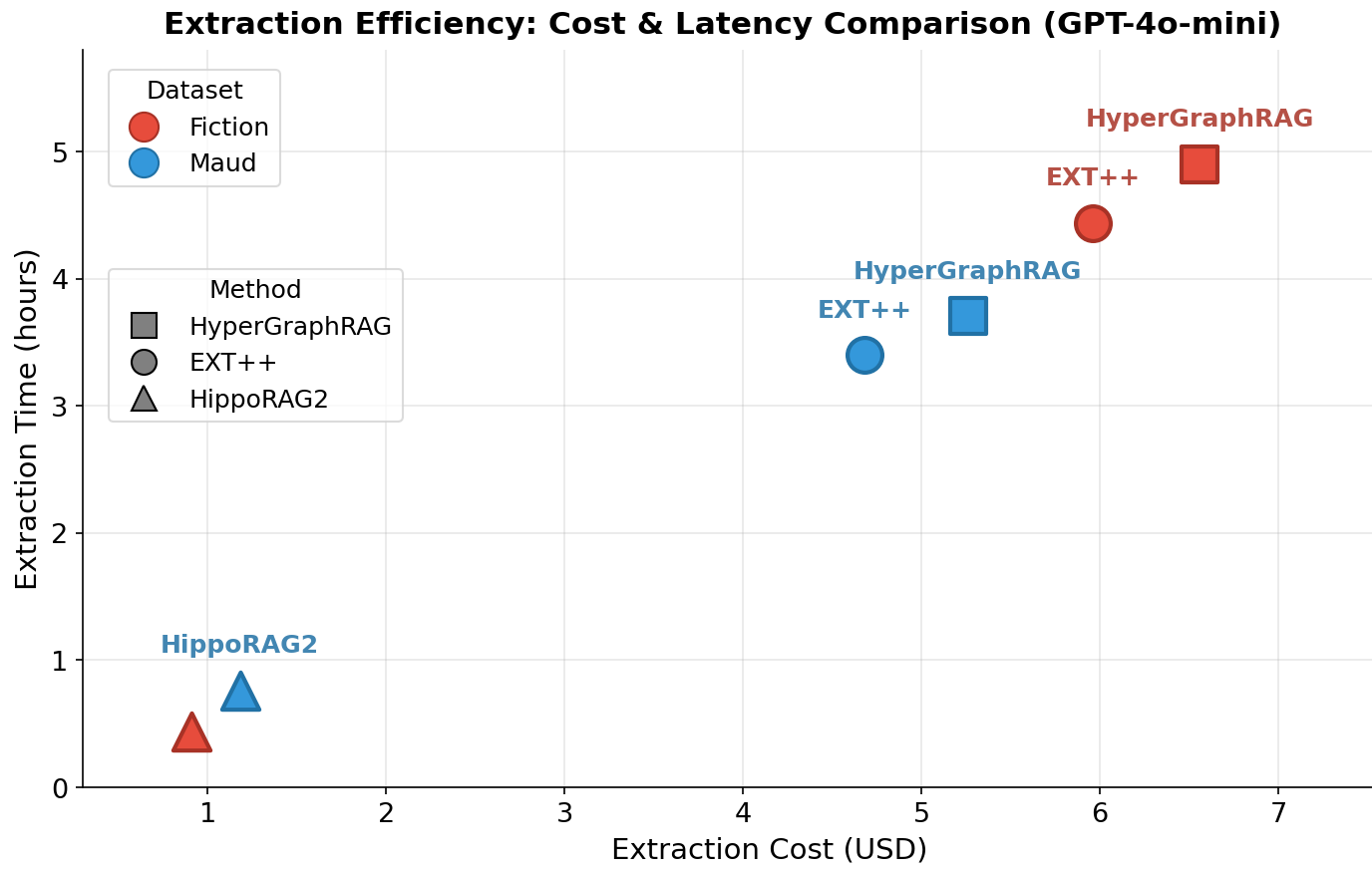}
    \caption{Extraction cost and latency comparison between HyperGraphRAG, EXT\textsuperscript{++}, and HippoRAG2}
    \label{fig:ext_cost}
\end{figure}

Figure~\ref{fig:ext_cost} compares the cost and total extraction latency between HyperGraphRAG (with and without EXT\textsuperscript{++}) and HippoRAG2 on the Fiction and MAUD datasets. Counter-intuitively, EXT\textsuperscript{++} --- which asks the model to perform three internal extraction passes before emitting the deduplicated union --- turns out to be faster and less expensive than simple extraction. This result is explained by two factors: (i) the longer EXT\textsuperscript{++} prompt (\textasciitilde600 additional tokens) benefits more from prefix caching (+20 to +24\% of cached tokens), reducing completion cost; (ii) the internal aggregation merges redundant extractions before generating output, reducing the number of completion tokens (-20\% on Fiction, -14\% on MAUD). Since LLM latency is dominated by autoregressive decoding, this reduction directly translates into shorter extraction time.

In contrast, HippoRAG2 exhibits significantly superior performance in terms of cost and latency, with a factor of approximately 5$\times$ to 7$\times$ depending on the metric. However, this efficiency stems directly from the nature of its representation: HippoRAG2 extracts a binary knowledge graph that is inherently less verbose than a hypergraph. This conciseness comes at a qualitative cost: binary graphs are known for their difficulty in faithfully representing temporal relations, epistemic modalities, and n-ary relations frequent in complex domains. The hypergraph, while more costly to construct, preserves the semantic richness of the source text and offers better robustness against these limitations.

\begin{figure}[ht]
    \centering
\includegraphics[width=0.5\textwidth]{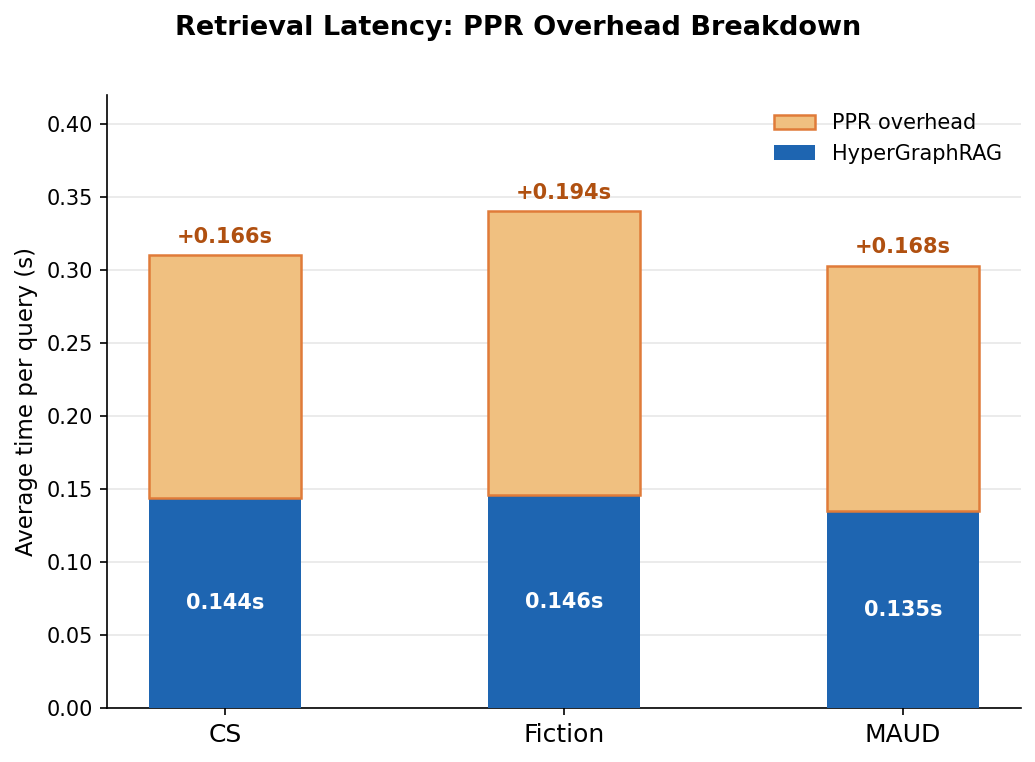}
    \caption{Average per-query retrieval latency with and without PPR}
    \label{fig:retrieval_time}
\end{figure}

\paragraph{Personalized PageRank latency:}
We evaluate the computational cost introduced by the Personalized PageRank (PPR) algorithm by comparing per-query retrieval latency with and without this optimization. The results in Figure~\ref{fig:retrieval_time} show remarkably low latency in both configurations. Without PPR, the average time per query ranges between 0.135s and 0.146s depending on the dataset. Enabling PPR incurs an overhead of less than 0.2 seconds per query across all tested corpora. This uniform behavior is explained by the fact that PPR operates on a local subgraph centered on the query-relevant entities, rather than on the entire knowledge hypergraph. This overhead remains negligible in a graph-augmented retrieval scenario, given the improvement in context quality provided to the model.
\vspace{-2mm}

\vspace{-3mm}
\section{Conclusion and Future Work}

This work extended the HyperGraphRAG approach through the integration of Personalized PageRank (PPR) and self-consistency extraction (EXT\textsuperscript{++}). PPR enables probabilistic relevance propagation through the hypergraph, significantly improving contextual recall and response completeness. EXT\textsuperscript{++} improves the graph's reliability upstream by reducing isolated hyperedges, with no additional latency or extraction cost. Evaluation on three corpora of distinct natures --- narrative, scientific, and legal --- confirms the superiority of our approach, with particularly marked gains in response correctness and completeness.

Several future directions are envisioned. First, diffusion methods natively designed for hypergraphs --- such as random walks on hypergraphs \cite{Yang2025} --- could better exploit the n-ary structure of hyperedges than standard PPR diffusion. Second, reconceiving each hyperedge as a condensed summary of atomic fragments rather than verbatim sentences would reduce the volume of produced tokens, thereby decreasing extraction cost and latency. Additionally, evaluating our approach on vector databases leveraging advanced search algorithms would enable measuring the impact on specialized single-hop benchmarks with redundant vocabulary, such as MAUD. Finally, a context \textit{pruning} mechanism would optimize token usage and reduce contextual noise before generation, paving the way for evaluation on other benchmarks requiring deeper reasoning or complex synthesis.

\bibliographystyle{plain}
\bibliography{biblio-ch-pfia}

@misc{luo2025hypergraphrag,
      title={HyperGraphRAG: Retrieval-Augmented Generation via Hypergraph-Structured Knowledge Representation}, 
      author={Haoran Luo and Haihong E and Guanting Chen and Yandan Zheng and Xiaobao Wu and Yikai Guo and Qika Lin and Yu Feng and Zemin Kuang and Meina Song and Yifan Zhu and Luu Anh Tuan},
      year={2025},
      eprint={2503.21322},
      archivePrefix={arXiv},
      primaryClass={cs.AI},
      note={NeurIPS 2025 poster},
      url={https://arxiv.org/abs/2503.21322}
}

@misc{anthropic2024contextual,
  author = {Anthropic},
  title = {Introducing Contextual Retrieval},
  year = {2024},
  note = {\url{https://www.anthropic.com/engineering/contextual-retrieval}},
  url = {https://www.anthropic.com/engineering/contextual-retrieval},
}

@misc{feng2025hyperrag,
      title={Hyper-RAG: Combating LLM Hallucinations using Hypergraph-Driven Retrieval-Augmented Generation}, 
      author={Yifan Feng and Hao Hu and Xingliang Hou and Shiquan Liu and Shihui Ying and Shaoyi Du and Han Hu and Yue Gao},
      year={2025},
      eprint={2504.08758},
      archivePrefix={arXiv},
      primaryClass={cs.IR},
      url={https://arxiv.org/abs/2504.08758}, 
}

@article{edge2024graphrag,
  title={From Local to Global: A Graph RAG Approach to Query-Focused Summarization},
  author={Edge, Darren and Trinh, Ha and Cheng, Yanlo and Bradley, Joshua and Chao, Alex and Mody, Apurva and Truitt, Shweti and Bansal, Gagan},
  journal={arXiv preprint arXiv:2404.16130},
  year={2024}
}

@misc{chen2025pathrag,
      title={PathRAG: Pruning Graph-based Retrieval Augmented Generation with Relational Paths}, 
      author={Boyu Chen and Zirui Guo and Zidan Yang and Yuluo Chen and Junze Chen and Zhenghao Liu and Chuan Shi and Cheng Yang},
      year={2025},
      eprint={2502.14902v2},
      archivePrefix={arXiv},
      primaryClass={cs.CL},
      url={https://arxiv.org/abs/2502.14902v2}, 
}

@inproceedings{
    gutierrez2025hipporag2,
    title={From {RAG} to Memory: Non-Parametric Continual Learning for Large Language Models},
    author={Bernal Jim{\'e}nez Guti{\'e}rrez and Yiheng Shu and Weijian Qi and Sizhe Zhou and Yu Su},
    booktitle={Forty-second International Conference on Machine Learning},
    year={2025},
    url={https://openreview.net/forum?id=LWH8yn4HS2}
}

@article{guo2024lightrag,
  title={LightRAG: Constructing and Utilizing Light-weight Knowledge Graph for Retrieval-Augmented Generation},
  author={Guo, Zirui and Fan, Zhaojun and Lu, Zhuoheng and Xu, Jun and Wen, Ji-Rong},
  journal={arXiv preprint arXiv:2410.05779},
  year={2024}
}

@article{zulun2026survey,
    author = {Zhu, Zulun and Huang, Tiancheng and Wang, Kai and Ye, Junda and Chen, Xinghe and Luo, Siqiang},
    title = {Graph-Based Approaches and Functionalities in Retrieval-Augmented Generation: A Comprehensive Survey},
    year = {2026},
    issn = {0360-0300},
    url = {https://doi.org/10.1145/3795880},
    doi = {10.1145/3795880},
    journal = {ACM Comput. Surv.},
}

@misc{zai2026proh,
      title={PRoH: Dynamic Planning and Reasoning over Knowledge Hypergraphs for Retrieval-Augmented Generation}, 
      author={Xiangjun Zai and Xingyu Tan and Xiaoyang Wang and Qing Liu and Xiwei Xu and Wenjie Zhang},
      year={2026},
      eprint={2510.12434},
      archivePrefix={arXiv},
      primaryClass={cs.CL},
      url={https://arxiv.org/abs/2510.12434}, 
}

@inproceedings{lei2025kag,
    author = {Liang, Lei and Bo, Zhongpu and Gui, Zhengke and Zhu, Zhongshu and Zhong, Ling and Zhao, Peilong and Sun, Mengshu and Zhang, Zhiqiang and Zhou, Jun and Chen, Wenguang and Zhang, Wen and Chen, Huajun},
    title = {KAG: Boosting LLMs in Professional Domains via Knowledge Augmented Generation},
    year = {2025},
    isbn = {9798400713316},
    doi = {10.1145/3701716.3715240},
    booktitle = {Companion Proceedings of the ACM on Web Conference 2025},
    pages = {334–343},
    location = {Sydney NSW, Australia},
    series = {WWW '25}
}

@inproceedings{he2024gretriever,
    title={G-Retriever: Retrieval-Augmented Generation for Textual Graph Understanding and Question Answering},
    author={Xiaoxin He and Yijun Tian and Yifei Sun and Nitesh V Chawla and Thomas Laurent and Yann LeCun and Xavier Bresson and Bryan Hooi},
    booktitle={The Thirty-eighth Annual Conference on Neural Information Processing Systems},
    year={2024},
    url={https://openreview.net/forum?id=MPJ3oXtTZl}
}

@inproceedings{hu-etal-2025-grag,
    title = "{GRAG}: Graph Retrieval-Augmented Generation",
    author = "Hu, Yuntong  and
      Lei, Zhihan  and
      Zhang, Zheng  and
      Pan, Bo  and
      Ling, Chen  and
      Zhao, Liang",
    booktitle = "Findings of the Association for Computational Linguistics: NAACL",
    year = {2025},
    url = "https://aclanthology.org/2025.findings-naacl.232/",
    doi = "10.18653/v1/2025.findings-naacl.232",
}

@inproceedings{yiqian2025ketrag,
author = {Huang, Yiqian and Zhang, Shiqi and Xiao, Xiaokui},
title = {KET-RAG: A Cost-Efficient Multi-Granular Indexing Framework for Graph-RAG},
year = {2025},
url = {https://doi.org/10.1145/3711896.3737012},
doi = {10.1145/3711896.3737012},
booktitle = {Proceedings of the 31st ACM SIGKDD Conference on Knowledge Discovery and Data Mining V.2},
location = {Toronto ON, Canada},
series = {KDD '25}
}

@inproceedings{lewis2020rag,
    author = {Lewis, Patrick and Perez, Ethan and Piktus, Aleksandra and Petroni, Fabio and Karpukhin, Vladimir and Goyal, Naman and K\"{u}ttler, Heinrich and Lewis, Mike and Yih, Wen-tau and Rockt\"{a}schel, Tim and Riedel, Sebastian and Kiela, Douwe},
    title = {Retrieval-augmented generation for knowledge-intensive NLP tasks},
    year = {2020},
    booktitle = {Proceedings of the 34th International Conference on Neural Information Processing Systems},
    articleno = {793},
    location = {Vancouver, BC, Canada},
}

@article{Xu2023LargeLM,
  title={Large language models for generative information extraction: a survey},
  author={Derong Xu and Wei Chen and Wenjun Peng and Chao Zhang and Tong Xu and Xiangyu Zhao and Xian Wu and Yefeng Zheng and Enhong Chen},
  journal={Frontiers of Computer Science},
  year={2024},
  volume={18},
  url={https://doi.org/10.1007/s11704-024-40555-y},
  doi = "10.1007/s11704-024-40555-y",
}

@inproceedings{chen2024universal,
    title={Universal Self-Consistency for Large Language Models},
    author={Xinyun Chen and Renat Aksitov and Uri Alon and Jie Ren and Kefan Xiao and Pengcheng Yin and Sushant Prakash and Charles Sutton and Xuezhi Wang and Denny Zhou},
    booktitle={ICML 2024 Workshop on In-Context Learning},
    year={2024},
    url={https://openreview.net/forum?id=LjsjHF7nAN}
}

@inproceedings{wang2023selfcons,
  author= {Xuezhi Wang and
                  Jason Wei and
                  Dale Schuurmans and
                  Quoc V. Le and
                  Ed H. Chi and
                  Sharan Narang and
                  Aakanksha Chowdhery and
                  Denny Zhou},
  title= {Self-Consistency Improves Chain of Thought Reasoning in Language Models},
  booktitle= {The Eleventh International Conference on Learning Representations,
                  {ICLR} 2023, Kigali, Rwanda},
  year={2023}
}

@article{pipitone2024legalbenchrag,
  title={LegalBench-RAG: A Benchmark for Retrieval-Augmented Generation in the Legal Domain},
  author={Pipitone, Nicholas and Houir Alami, Ghita},
  journal={arXiv preprint arXiv:2408.10343},
  year={2024},
  url={https://arxiv.org/abs/2408.10343}
}

@inproceedings{qian2025memorag,
  title     = {MemoRAG: Boosting Long Context Processing with Global Memory-Enhanced Retrieval Augmentation},
  author    = {Hongjin Qian and Zheng Liu and Peitian Zhang and Kelong Mao and Defu Lian and Zhicheng Dou and Tiejun Huang},
  booktitle = {Proceedings of the ACM Web Conference)},
  year = {2025},
  url       = {https://arxiv.org/abs/2409.05591},
}

@misc{neo4j_graphtransformer,
  title        = {Creating knowledge graphs from unstructured data},
  author       = {{Neo4j}},
  howpublished = {\url{https://neo4j.com/developer/genai-ecosystem/importing-graph-from-unstructured-data/}},
}

@inproceedings{es-etal-2024-ragas,
    title = "{RAGA}s: Automated Evaluation of Retrieval Augmented Generation",
    author = "Es, Shahul  and
      James, Jithin  and
      Espinosa Anke, Luis  and
      Schockaert, Steven",
    editor = "Aletras, Nikolaos  and
      De Clercq, Orphee",
    booktitle = "Proceedings of the 18th Conference of the European Chapter of the Association for Computational Linguistics: System Demonstrations",
    year = "2024",
    url = "https://aclanthology.org/2024.eacl-demo.16/",
}

@article{Yang2025,
  author    = {Mu-Rong Yang and Xin-Jian Xu},
  title     = {Recent Advances in Hypergraph Neural Networks},
  journal   = {Journal of the Operations Research Society of China},
  year      = {2025},
  doi       = {10.1007/s40305-025-00630-y},
  url       = {https://doi.org/10.1007/s40305-025-00630-y},
  issn      = {2194-6698},
}

\newpage
\appendix
\section{Prompts}
\label{appendix:prompt}

\begin{strip}
Prompt~1 presents the instructions used by the EXT\textsuperscript{++} method for optimized knowledge hypergraph extraction and Prompt~2 presents the LLM-as-judge evaluator. 

\begin{tcolorbox}[
  colback=gray!10,
  colframe=black,
  width=\textwidth
]
\small
Given a text document, identify all entities and relationships present within the text.

\textbf{\# Self-Consistency Protocol}

\begin{enumerate}
  \item Extraction Runs:
  \begin{itemize}[leftmargin=1.5em]
    \item Internally perform entities and knowledge segments extraction three times independently.
    \item Each run must complete all steps as a separate, uninfluenced process.
  \end{itemize}

  \item Aggregation and Deduplication after completing all extraction runs:
  \begin{itemize}[leftmargin=1.5em]
  \item For Knowledge Segments (hyper-relations):
  \begin{itemize}[leftmargin=1em]
    \item Merge semantically similar or overlapping knowledge segments into a single, comprehensive segment.
    \item If multiple segments describe the same concept/relationship, combine them into ONE segment with the highest completeness score.
  \end{itemize}

  \item For Entities:
  \begin{itemize}[leftmargin=1em]
    \item Merge entities that refer to the same concept even if worded differently and choose the most specific entity name.
    \item If contradictions are found, resolve in favor of the most consistent information.
  \end{itemize}
    \end{itemize}
\end{enumerate}

\textbf{\# Extraction Steps}

\begin{enumerate}
  \item Divide the text into complete knowledge segments (hyper-relations). For each knowledge segment, extract the following information:
  \begin{itemize}[leftmargin=1.5em]
    \item \texttt{knowledge\_segment}: Extract 1--3 consecutive sentences EXACTLY as they appear in the source text. DO NOT paraphrase, modify, or rewrite the text. The extracted span should be semantically complete and self-contained.
    \item \texttt{completeness\_score}: A score from 0 to 10 indicating the completeness of the knowledge segment.
  \end{itemize}

  \item For each knowledge segment, identify all related entities. For each identified entity, extract the following information:
  \begin{itemize}[leftmargin=1.5em]
    \item \texttt{entity\_name}: Name of the entity. Resolve coreferences based on the whole context and provide consistent naming.
    \item \texttt{entity\_type}: Type of the entity.
    \item \texttt{entity\_description}: Comprehensive description of the entity's attributes and activities.
  \end{itemize}
\end{enumerate}
Output each knowledge segment followed immediately by its associated entities.
\end{tcolorbox}
\vspace{-2mm}
\promptcaption{Self-consistency extraction prompt (EXT\textsuperscript{++}) for knowledge hypergraph construction}

\vspace{2mm}
\begin{tcolorbox}[
  colback=gray!10,
  colframe=black,
  width=\textwidth
]
\small
You are an expert evaluator assessing the quality of RAG-generated answers. You must consider both the reference answer and the original question as sources of key information points to evaluate the RAG answer. You must not penalize the RAG answer for additional information, unless it is incorrect or misleading. Evaluate the following dimensions:

\vspace{0.5em}
\textbf{\# COMPLETENESS: } Whether RAG answer considers all important aspects and is thorough using the following scoring guide:
\begin{itemize}
  \item \textbf{10}: Covers ALL key information from the reference answer even if the wording or structure differs. 
  \item \textbf{8--9}: Covers most key information but misses 1--2 minor details or supporting facts.
  \item \textbf{6--7}: Covers some important aspects, but lacks depth or overlooks notable areas.
  \item \textbf{4--5}: Touches on a few relevant points, but overall lacks substance or completeness.
  \item \textbf{1--3}: Sparse or shallow treatment of the topic; misses most key aspects.
  \item \textbf{0}: No comprehensiveness at all; completely superficial or irrelevant.
\end{itemize}
\textbf{\# CORRECTNESS} Whether RAG answer is logically and factually correct using the following scoring guide:
\begin{itemize}
  \item \textbf{10}: Fully accurate and logically sound; no flaws in reasoning or facts.
  \item \textbf{8--9}: Mostly correct with minor inaccuracies or small logical gaps.
  \item \textbf{6--7}: Partially correct; some key flaws or inconsistencies present.
  \item \textbf{4--5}: Noticeable incorrect reasoning or factual errors throughout.
  \item \textbf{1--3}: Largely incorrect, misleading, or illogical.
  \item \textbf{0}: Entirely wrong or nonsensical.
\end{itemize}
\end{tcolorbox}
\vspace{-2mm}
\promptcaption{LLM-as-judge evaluation prompt for Correctness and Completeness scoring}
\end{strip}

\onecolumn
\section{Use Case}

Table~\ref{tab:usecase} presents a qualitative comparison on a multi-hop fiction query ``How does Vadassy's plan to trap the spy ultimately fail?'' requiring the synthesis of three narrative passages. The golden answer identifies three key facts: Vadassy is \emph{locked in the writing-room}, the spy \emph{retrieves the camera without being identified}, and Vadassy is left \emph{without evidence}.

\begin{table*}[ht]
\centering
\small
\caption{Qualitative comparison of generated answers for a multi-hop fiction query. \textbf{Bold} text highlights key facts; correctness and completeness are scored from 0 to 10.}
\label{tab:usecase}
\renewcommand{\arraystretch}{1.3}
\begin{tabularx}{\textwidth}{|>{\bfseries\RaggedRight}p{3.2cm}|>{\RaggedRight\arraybackslash}X|c|c|}
\hline
\textbf{} & \textbf{How does Vadassy's plan to trap the spy ultimately fail?} & \textbf{Correctness} & \textbf{Completeness} \\
\hline
Golden Answer &
Vadassy's plan to trap the spy fails when he is \textbf{locked in the writing-room} by the person who takes his camera, which was left as bait. This person successfully \textbf{retrieves the camera without being identified}, leaving Vadassy \textbf{without evidence} and in a \textbf{compromised position}.
& -- & -- \\
\hline
HippoRAG2 &
Vadassy's plan fails because he underestimates the spy's intelligence and the police's strategy, leading to the \textbf{spy's escape} before he can be apprehended.
& 5 & 3 \\
\hline
HyperGraphRAG &
Vadassy's plan to trap the spy ultimately fails because he was actually operating within a larger scheme orchestrated by Beghin. While Vadassy believed he was cleverly setting a trap for the spy, his efforts were futile since the authorities already knew the spy's identity and could have arrested them at any time. The plan backfired spectacularly when \textbf{Vadassy himself was arrested} on espionage charges in front of all the hotel guests, effectively \textbf{becoming caught in his own trap} rather than catching the real spy. His attempts at subterfuge proved ineffective because he was unknowingly being used as a pawn in Beghin's broader strategy to pressure the actual spy.
& 10 & 8 \\
\hline
HyperGraphRAG + PPR &
Vadassy's plan to trap the spy fails in multiple ways. His attempts to alarm the spy through \textbf{camera inquiries} and reporting a fake robbery backfire due to his inexperience and limited understanding of the situation. The spy proves more cunning, managing to search Vadassy's room and \textbf{steal his film rolls} instead. While Vadassy tries to investigate hotel guests, K\"{o}che quickly sees through his amateur detective work. The ultimate irony is that \textbf{Vadassy's trap catches himself} --- he ends up being arrested at the R\'{e}serve while \textbf{the spy remains free}. His failure stems from both his inexperience in espionage matters and the fact that he was unknowingly being manipulated as part of a larger police operation, where Beghin and others already knew the spy's identity but kept Vadassy in the dark.
& 10 & 9 \\
\hline
HyperGraphRAG + EXT\textsuperscript{++} + PPR &
Vadassy's plan to trap the spy failed because while he was watching the \textbf{planted camera} through a mirror in the writing room, the spy outsmarted him by approaching from behind, \textbf{locking him in the room}, and \textbf{stealing the camera} while he was trapped. By the time Vadassy escaped through the window, both \textbf{the spy and the camera were gone}, leaving him \textbf{without the evidence} he needed to prove his innocence. Rather than catching the spy, \textbf{Vadassy himself became the one who was trapped.}
& 10 & 10 \\
\hline
\end{tabularx}
\end{table*}

HippoRAG2 produces a vague and generic response, mentioning only ``the spy's escape'' without recovering any of the specific mechanisms described in the reference. This illustrates the inability of binary graph retrieval to gather the dispersed narrative fragments needed for multi-hop reasoning. HyperGraphRAG captures the broader narrative arc --- Vadassy being caught in his own trap --- but omits the precise mechanism (the writing-room locking and camera theft). Adding PPR enriches the retrieved context: the answer now mentions the camera inquiries and Vadassy's self-entrapment, yet still lacks the specific writing-room detail. Our full approach, HyperGraphRAG + EXT\textsuperscript{++} + PPR, recovers \emph{all} key facts demonstrating how the combination of a structurally richer hypergraph and global PPR propagation enables exhaustive multi-hop retrieval.

Notably, correctness scores are identical (10/10) for the three hypergraph-based approaches. This is because the correctness metric evaluates whether the generated answer contains factual errors relative to the reference; since these three responses do not introduce false statements, they all receive the maximum score. Correctness is thus insensitive to \emph{missing} information --- an answer can be perfectly correct yet incomplete. The completeness metric, by contrast, specifically penalizes omissions, making it the decisive discriminator for multi-hop queries where coverage of all relevant narrative elements is critical.

\end{document}